\pdfoutput=1

\documentclass[11pt]{article}

\usepackage{ACL2023}

\usepackage{times}
\usepackage{latexsym}

\usepackage[T1]{fontenc}
\usepackage[utf8]{inputenc}

\usepackage{microtype}
\usepackage{inconsolata}

\usepackage{graphicx}       
\usepackage{amsmath}        
\usepackage{amssymb}
\usepackage{booktabs}       
\usepackage{multirow}
\usepackage{mathtools}      
\usepackage{filecontents}   
\usepackage{adjustbox}      
\usepackage{siunitx}        

\title{DyKen-Hyena: Dynamic Kernel Generation via Cross-Modal Attention for Multimodal Intent Recognition}

\author{
  Yifei Wang\textsuperscript{1}, \
  Wenbin Wang\textsuperscript{1}, \
  Yong Luo\textsuperscript{1,} \thanks{Yong Luo is the corresponding author.} 
  \\[1ex]
  \textsuperscript{1}Wuhan University \\
}

\begin{document}
\maketitle
\begin{abstract}
Though Multimodal Intent Recognition (MIR) proves effective by utilizing rich information from multiple sources (e.g., language, video, and audio), the potential for intent-irrelevant and conflicting information across modalities may hinder performance from being further improved. Most current models attempt to fuse modalities by applying mechanisms like multi-head attention to unimodal feature sequences and then adding the result back to the original representation. This process risks corrupting the primary linguistic features with noisy or irrelevant non-verbal signals, as it often fails to capture the fine-grained, token-level influence where non-verbal cues should modulate, not just augment, textual meaning.
To address this, we introduce DyKen-Hyena, which reframes the problem from feature fusion to processing modulation. Our model translates audio-visual cues into dynamic, per-token convolutional kernels that directly modulate textual feature extraction. This fine-grained approach achieves state-of-the-art results on the MIntRec and MIntRec2.0 benchmarks. Notably, it yields a +10.46\% F1-score improvement in out-of-scope detection, validating that our method creates a fundamentally more robust intent representation. 
\end{abstract}

\section{Introduction}
The phrase "that's great" can express genuine enthusiasm, biting sarcasm, or hesitant doubt. The literal text is identical; the true intent is conveyed almost entirely through the rich, non-verbal context of vocal intonation and facial expression. This inherent ambiguity of language lies at the heart of the challenge for Multimodal Intent Recognition, a critical capability for next-generation conversational AI \cite{yu2021mintrec, shenoy2024mintrec2}. As a key subfield of Multimodal Machine Learning and Affective Computing \cite{baltruvsaitis2018multimodal, poria2020beneath}, mastering this capability is essential for applications ranging from empathetic healthcare assistants to intuitive in-car control systems.
Prevailing research has largely framed this as a problem of \textit{feature fusion} at the representation level. A dominant strategy involves using powerful unimodal encoders to independently process text, audio, and video streams into high-level representations. These representations are then fused at an intermediate stage, often through sophisticated mechanisms. Whether through cross-modal attention \cite{tsai2019multimodal}, gating mechanisms \cite{rahman2020integrating}, or even within unified Transformer backbones \cite{tan2019lxmert}, the fusion act typically occurs between pre-digested feature vectors. While effective, this paradigm suffers from a fundamental limitation: it attempts to reconcile modalities only after their low-level temporal dynamics and momentary interactions have been abstracted away. This is akin to reading a transcript of a conversation while only getting periodic summaries of the emotional tone, rather than hearing how the tone of voice colors each specific word.
In this paper, we advocate for a paradigm shift from feature fusion to processing modulation. Instead of asking how to \textit{combine} features, we ask: how can non-verbal cues dynamically \textit{steer} the process of language understanding itself? We propose that multimodal fusion should be an early, integral, and modulatory act, not a late-stage combination. To realize this vision, we introduce \textbf{DyKen-Hyena} (Dynamic Kernel Hyena), a novel architecture designed for deep, context-aware fusion at the earliest point of representation learning.
The core principle of DyKen-Hyena is to use non-verbal signals to determine \textit{how} each text token should be computationally processed. Our model first employs a cross-modal attention mechanism, where each text token's embedding "queries" the corresponding audio and visual streams to produce a token-specific context vector. This vector is then passed through a lightweight generator network to produce a unique, per-token short convolutional kernel. These dynamically generated kernels, embodying the real-time audio-visual context, are integrated into a powerful and efficient Hyena sequence model \cite{poli2023hyena} to process the initial text embeddings. By embedding this entire fusion module at the input stage of a BERT encoder \cite{devlin2019bert}, we ensure that this fine-grained, multimodally-informed representation propagates through all subsequent layers of deep contextualization.
We validate our approach through extensive experiments on the MIntRec \cite{yu2021mintrec} and the larger, more complex MIntRec2.0 \cite{shenoy2024mintrec2} benchmarks. Our contributions are demonstrated by DyKen-Hyena's state-of-the-art performance. Most notably, its success in recognizing intents heavily reliant on non-verbal cues (e.g., "Joke," "Taunt") and its remarkable +10.46\% absolute F1-score improvement in out-of-scope detection underscore the robustness of our approach. This indicates that by learning a more fundamentally sound representation of multimodal signals, our model is not only more accurate for known intents but also significantly more adept at identifying unknown ones. Our work represents a step towards AI that understands not just what is said, but how it is said.

\begin{figure*}[!t]
    \centering
    \includegraphics[width=\textwidth]{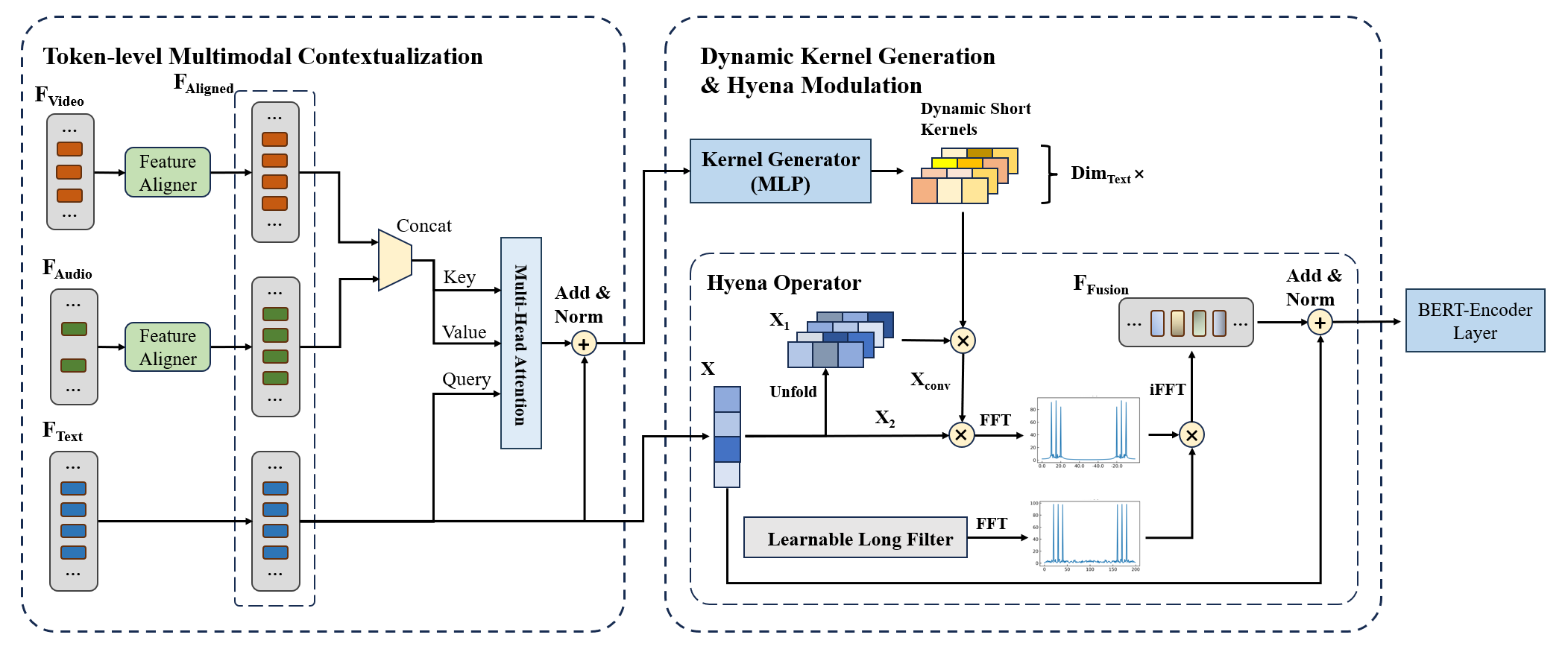} 
    \caption{The overall architecture of our DyKen-Hyena framework. The model consists of two main stages. (1) \textbf{Token-level Multimodal Contextualization}: For each text token embedding (Query), a cross-modal attention mechanism attends to aligned audio and visual feature sequences (Key/Value) to produce a token-aligned multimodal context. (2) \textbf{Dynamic Kernel Generation \& Hyena Modulation}: This context is fed into a Kernel Generator (a lightweight MLP) to create unique short convolution kernels for each token. These dynamic kernels, along with a static long-range filter, are used by the Hyena Operator to modulate the text features. The resulting deeply fused representation is then passed to the main BERT encoder.}
    \label{fig:overall_architecture}
\end{figure*}

\section{Related Work}
Our work is positioned at the intersection of three key research areas: multimodal fusion strategies, the application of Transformers in multimodal contexts, and the emerging field of efficient, sub-quadratic sequence models and dynamic networks.
\subsection{Multimodal Fusion Strategies}
The central challenge in multimodal learning lies in effectively integrating information from heterogeneous data streams. A comprehensive survey by Baltrusaitis et al. \cite{baltruvsaitis2018multimodal} categorizes fusion strategies into early, late, and hybrid approaches. Early fusion, which concatenates raw or low-level features at the input stage \cite{poria2017context}, allows the model to learn cross-modal interactions from the outset but can be susceptible to data asynchrony and differing feature granularities. Conversely, late fusion combines the outputs of powerful unimodal models at the decision level, offering robustness but failing to capture the fine-grained, low-level interactions crucial for tasks like intent recognition \cite{nojavanasghari2016deep}.
Most contemporary research resides in hybrid fusion, which integrates modalities at intermediate network layers. This area has been dominated by several key paradigms:
\begin{itemize}
\item \textbf{Attention-Based Fusion:} Cross-modal attention, popularized by models like MulT \cite{tsai2019multimodal}, has become a de facto standard. It allows one modality to dynamically attend to relevant features in another, effectively aligning and weighing information. This principle forms the backbone of numerous vision-language models \cite{lu2019vilbert, tan2019lxmert}.
\item \textbf{Tensor-Based Fusion:} Methods like the Tensor Fusion Network (TFN) \cite{zadeh2017tensor} explicitly model unimodal, bimodal, and trimodal interactions through a tensor outer product. While powerful in capturing high-order correlations, their computational complexity and parameter count can be prohibitive. Subsequent work, such as Low-rank Multimodal Fusion (LMF) \cite{liu2018efficient}, has sought to mitigate this cost through low-rank tensor approximations.
\item \textbf{Gating Mechanisms:} Models such as MAG-BERT \cite{rahman2020integrating} employ gating units that learn to control the flow of information from each modality. This data-driven mechanism dynamically adjusts the contribution of each modality, preventing any single one from overwhelming the others.
\end{itemize}
Our approach, DyKen-Hyena, introduces a new perspective on hybrid fusion. Instead of fusing pre-computed feature vectors through attention or gating, we propose a \textit{modulatory fusion}. Non-verbal signals are not just another source of features to be merged; they become directives that actively shape and control the computational process applied to the textual modality at the feature extraction level. This represents a fundamental shift from combining what is seen and heard to dynamically altering how what is read is understood.
\subsection{Transformers in Multimodal Learning}
The advent of the Transformer \cite{vaswani2017attention} and pre-trained models like BERT \cite{devlin2019bert} revolutionized NLP and quickly permeated multimodal research. Early adaptations like MulT \cite{tsai2019multimodal} applied the cross-attention mechanism to fuse sequences. However, a major paradigm shift occurred with the development of large-scale, pre-trained multimodal Transformers. Models can be broadly categorized as two-stream, like ViLBERT \cite{lu2019vilbert} and LXMERT \cite{tan2019lxmert}, which use separate Transformers to encode vision and language before fusing them through co-attention layers; or single-stream, like UNITER \cite{chen2020uniter} and Oscar \cite{li2020oscar}, which concatenate visual and textual tokens into a single sequence processed by a unified Transformer. More recently, models like Data2Vec \cite{baevski2022data2vec} have pushed towards modality-agnostic pre-training using a single architecture.
Our work leverages the power of a pre-trained BERT encoder for high-level contextualization. However, we contend that the quality of the input representation fed to such encoders is paramount. DyKen-Hyena acts as a sophisticated, multimodally-aware "embedding processor" that is placed before the main BERT encoder. By deeply fusing and modulating the text features at the input level, we provide the Transformer with a representation that is already rich with cross-modal context, allowing its self-attention layers to focus on higher-order semantic relationships from a much more informed starting point.
\subsection{Efficient Sequence Models and Dynamic Networks}
The quadratic complexity of self-attention in Transformers has spurred the development of more efficient architectures for long-sequence modeling. A prominent line of research has revisited convolutions and state-space models (SSMs). The Hyena operator \cite{poli2023hyena}, which our work adapts, is a key example from the convolution-based family. It replaces attention with a combination of a long convolution, implemented efficiently via FFT, and data-controlled gating to achieve near-linear complexity while maintaining a global receptive field.
Parallel to this, State Space Models (SSMs) have emerged as a powerful alternative. Models like the Structured State Space Sequence Model (S4) \cite{gu2021efficiently} demonstrated the potential of continuous-time systems for sequence modeling. More recently, Mamba \cite{gu2023mamba} introduced a selection mechanism that allows SSM parameters to be input-dependent, achieving state-of-the-art performance on various modalities with true linear-time complexity.
Our work draws inspiration from two key concepts here. First, we adopt the efficient convolutional structure of Hyena. Second, and more critically, we connect this to the principle of \textbf{dynamic networks}, where network parameters are generated on-the-fly conditioned on the input \cite{jia2016dynamic}. While Mamba makes its internal state transitions dependent on the input sequence itself, our innovation is to make the feature extraction operator—the short convolutional kernel—dynamic and conditional on \textit{external, cross-modal signals}. By having audio-visual cues generate the specific processing kernels for each text token, we create a fusion mechanism that is not only contextually nuanced and highly adaptive but also computationally efficient. This novel synthesis of efficient sequence modeling and cross-modal dynamic computation is the core contribution of our work.

\section{Methodology}
Our framework, illustrated in Figure~\ref{fig:overall_architecture}, is built upon a BERT-based architecture. We innovate by inserting a \textit{Dynamic Kernel Generation and Hyena Modulation} module between the initial text embedding layer and the main BERT encoder, enabling deep fusion before the high-level contextualization process begins.

\subsection{Token-level Multimodal Contextualization}
The first step is to create a contextualized representation for each text token that is richly informed by the audio and visual modalities. Given text features $\mathbf{F}_t \in \mathbb{R}^{L_t \times D_t}$, audio features $\mathbf{F}_a \in \mathbb{R}^{L_a \times D_a}$, and visual features $\mathbf{F}_v \in \mathbb{R}^{L_v \times D_v}$, we first align the audio and visual sequences to the text sequence length $L_t$ using standard alignment techniques.

The aligned non-textual features are projected to the dimension $D_t$ and concatenated to form a unified non-textual context $\mathbf{F}_{av} \in \mathbb{R}^{L_t \times 2D_t}$. We then employ a multi-head attention mechanism where the text tokens act as queries to "attend to" the most relevant audio-visual cues at each time step:

\begin{equation}
\mathbf{C}_{attn} = \text{MultiHeadAttn}\Big(
\begin{aligned}
    &\mathbf{Q}=\mathbf{F}_t, \\
    &\mathbf{K}=\mathbf{F}_{av}, \\
    &\mathbf{V}=\mathbf{F}_{av}
\end{aligned}
\Big)
\end{equation}

The output, $\mathbf{C}_{attn} \in \mathbb{R}^{L_t \times D_t}$, represents the aggregated audio-visual context for each word. A residual connection and layer normalization are applied to stabilize training and integrate this context with the original text information, yielding the final local context $\mathbf{C}_{local}$:
\begin{equation}
    \mathbf{C}_{local} = \text{LayerNorm}(\mathbf{C}_{attn} + \mathbf{F}_t)
\end{equation}

\subsection{Dynamic Kernel Generation}
The token-specific multimodal context $\mathbf{C}_{local}$ now holds the necessary information to guide text processing. It is fed into a small MLP, the \textit{Kernel Generator}, to produce the parameters of the short convolution kernels:
\begin{equation}
    \mathbf{K}_{params} = \text{MLP}(\mathbf{C}_{local}) \in \mathbb{R}^{L_t \times (D_t \cdot K_s)}
\end{equation}
where $K_s$ is the desired short kernel size (e.g., 3). These parameters are then dynamically reshaped to form the per-token short convolution filters $\mathbf{F}_{short} \in \mathbb{R}^{L_t \times D_t \times K_s}$. This step is the crux of our approach: we have translated non-verbal information into a set of processing operators (kernels) tailored for each word.

\subsection{Hyena Operator with Dynamic Modulation}
We adapt the powerful Hyena operator \cite{poli2023hyena} to incorporate our dynamic kernels. The original text embeddings $\mathbf{F}_t$ serve as the primary input, which are projected and split into two parallel processing paths, $\mathbf{X}_1$ and $\mathbf{X}_2$.

\paragraph{Dynamic Short Convolution.} To apply our per-token kernels efficiently, we first use an `unfold` operation to create a sliding-window view of $\mathbf{X}_1 \in \mathbb{R}^{L_t \times D_t}$. The local convolution is then efficiently computed via an element-wise product between this unfolded view and our dynamic kernel bank $\mathbf{F}_{short}$, resulting in the locally modulated representation $\mathbf{X}_{conv}$. This operation allows, for example, the kernel generated from a sarcastic tone to sharpen or invert the features of the word "great".

\paragraph{Long Convolution and Gating.} The output $\mathbf{X}_{conv}$ is gated element-wise with the parallel path $\mathbf{X}_2$. This data-dependent gating mechanism controls the flow of information. The gated result is then processed by a long convolution, implemented efficiently via Fast Fourier Transform (FFT), which captures global, long-range dependencies within the sequence. The final output of the Hyena operator is denoted $\mathbf{F}_{fusion}$.

\subsection{Integration with BERT Encoder}
The entire fusion module is applied to the initial word embeddings. The resulting deeply fused embeddings $\mathbf{F}_{fused}$ are produced via a final residual connection with the original text input $\mathbf{F}_t$:
\begin{equation}
    \mathbf{F}_{final} = \text{LayerNorm}(\mathbf{F}_{fusion} + \mathbf{F}_t)
\end{equation}
These fused embeddings, now rich with multimodal context, are passed to the standard BERT encoder layers for high-level contextual representation learning.

\section{Experiments}
\subsection{Datasets and Metrics}
We evaluate our model on two public benchmarks: \textbf{MIntRec} \cite{yu2021mintrec} and the larger, more challenging \textbf{MIntRec2.0} \cite{shenoy2024mintrec2}, which includes out-of-scope (OOS) intent detection. We report standard classification metrics: Accuracy (Acc), Weighted F1 (WF), Weighted Precision (WP), Macro F1 (F1), Precision (Prec), and Recall (Rec). For MIntRec2.0, following the official benchmark, we also report OID Accuracy (oid\_acc), F1 for in-scope (F1-IS) and out-of-scope (F1-OOS) intents, and OID F1 (oid\_f1).

\subsection{Baselines}
We compare DyKen-Hyena against two strong and widely recognized baselines that represent different fusion strategies:
\begin{itemize}
    \item \textbf{MulT} \cite{tsai2019multimodal}: A classic Transformer-based model using cross-modal attention for fusion.
    \item \textbf{MAG-BERT} \cite{rahman2020integrating}: A strong baseline that uses a gating mechanism to fuse modalities with a pre-trained BERT model.
\end{itemize}

\subsection{Implementation Details}
Our framework uses a `bert-base-uncased` model as its backbone. All reported results are the average of 5 runs with different random seeds to ensure statistical significance. Our primary model variant is DyKen-Hyena with a dynamic kernel size of $K_s=3$.

\begin{table*}[!t]
\centering
\small
\begin{adjustbox}{width=\textwidth,center}
\begin{tabular}{l|S[table-format=2.2] S[table-format=2.2] S[table-format=2.2] S[table-format=2.2] S[table-format=2.2] S[table-format=2.2]|S[table-format=2.2] S[table-format=2.2] S[table-format=2.2] S[table-format=2.2]}
\toprule
\textbf{Model} & {\textbf{Acc}} & {\textbf{WF}} & {\textbf{WP}} & {\textbf{F1}} & {\textbf{Prec}} & {\textbf{Rec}} & {\textbf{oid\_acc}} & {\textbf{F1-IS}} & {\textbf{F1-OOS}} & {\textbf{oid\_f1}} \\
\midrule
MulT \cite{tsai2019multimodal} & 60.20 & 58.76 & 59.46 & 52.63 & 57.03 & 52.58 & 42.48 & 44.24 & 25.18 & 43.62 \\
MAG \cite{rahman2020integrating} & \bfseries \textbf{60.62} & \underline{59.53} & \underline{59.80} & \underline{54.17} & \underline{57.36} & \underline{54.01} & \underline{43.51} & \underline{45.62} & \underline{28.23} & \underline{45.06} \\
\midrule
\textbf{DyKen-Hyena (Ours)} & \underline{60.54} & \bfseries \textbf{59.79} & \bfseries \textbf{60.21} & \bfseries \textbf{55.02} & \bfseries \textbf{58.35} & \bfseries \textbf{54.53} & \bfseries \textbf{45.63} & \bfseries \textbf{46.21} & \bfseries \textbf{38.69} & \bfseries \textbf{45.97} \\
\midrule
\textit{$\Delta$} & \textit{-0.08} & \textit{+0.26} & \textit{+0.41} & \textit{+0.85} & \textit{+0.99} & \textit{+0.52} & \textit{+2.12} & \textit{+0.59} & \textit{+10.46} & \textit{+0.91} \\
\bottomrule
\end{tabular}
\end{adjustbox}
\caption{Main performance comparison on the MIntRec2.0 test set (\%). Best results are in \textbf{bold}, second best are \underline{underlined}. $\Delta$ is the improvement of our model over the best baseline in that column. Our model excels in robust metrics like Macro F1 and especially in open-set detection (F1-OOS).}
\label{tab:main_results_mintrec2}
\end{table*}

\begin{table*}[!t]
\centering
\small
\begin{tabular}{l|S[table-format=2.2] S[table-format=2.2] S[table-format=2.2] S[table-format=2.2] S[table-format=2.2] S[table-format=2.2]}
\toprule
\textbf{Model} & {\textbf{Acc}} & {\textbf{F1}} & {\textbf{Prec}} & {\textbf{Rec}} & {\textbf{WF}} & {\textbf{WP}} \\
\midrule
MulT \cite{tsai2019multimodal} & 71.78 & \underline{68.47} & \underline{69.73} & 68.27 & \underline{71.58} & \underline{72.15} \\
MAG \cite{rahman2020integrating}  & \underline{71.82} & 67.92 & 68.37 & \underline{68.56} & 71.53 & 72.08 \\
\midrule
\textbf{DyKen-Hyena (Ours)} & \bfseries \textbf{73.66} & \bfseries \textbf{69.26} & \bfseries \textbf{70.30} & \bfseries \textbf{69.54} & \bfseries \textbf{73.05} & \bfseries \textbf{73.30} \\
\midrule
\textit{$\Delta$} & \textit{+1.84} & \textit{+0.79} & \textit{+0.57} & \textit{+0.98} & \textit{+1.47} & \textit{+1.15} \\
\bottomrule
\end{tabular}
\caption{Performance comparison on the MIntRec test set (\%). Best results are in \textbf{bold}, second best are \underline{underlined}. Our model achieves significant improvements across all major metrics.}
\label{tab:main_results_mintrec1}
\end{table*}

\section{Results and Analysis}
\subsection{Main Results}
As shown in Tables \ref{tab:main_results_mintrec2} and \ref{tab:main_results_mintrec1}, DyKen-Hyena demonstrates superior or highly competitive performance across the board, setting a new state-of-the-art on both benchmarks.

On the MIntRec dataset (Table \ref{tab:main_results_mintrec1}), which focuses on closed-set classification, DyKen-Hyena achieves pronounced gains over the strong baselines. It surpasses the best prior result by a substantial margin of \textbf{+1.84\%} in Accuracy and \textbf{+1.47\%} in Weighted F1, confirming the overall effectiveness and superiority of our dynamic fusion approach in a standard setting.

On the more complex MIntRec2.0 dataset (Table \ref{tab:main_results_mintrec2}), which includes an out-of-scope detection task, the results tell a more nuanced and compelling story. While the overall accuracy is comparable to the best baseline, our model shows significant advantages in more robust and challenging metrics. Notably, Macro F1, which treats all classes equally and is less sensitive to class imbalance, improves by \textbf{+0.85\%}. The most striking improvement is in out-of-scope detection, where our DyKen-Hyena model achieves an F1-OOS score of \textbf{38.69\%}. This represents a massive absolute increase of \textbf{+10.46\%} over the best baseline, indicating a fundamentally better ability to distinguish between known and unknown intents. This suggests our model learns a tighter, more accurate representation of in-scope intents, making it more robust to novel inputs.

\begin{figure*}[!t]
    \centering
    \includegraphics[width=\textwidth]{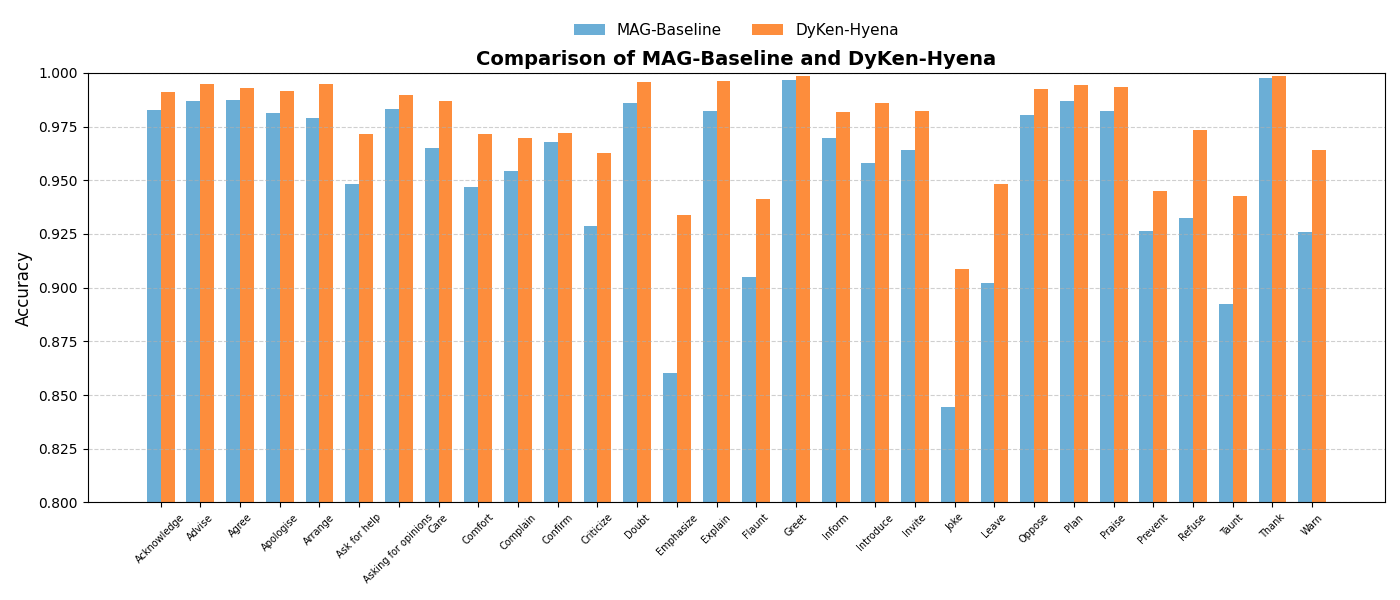}
    \caption{Per-class accuracy comparison between MAG-BERT and our DyKen-Hyena on the MIntRec2.0 dataset. Our model shows significant improvements on intents that are highly ambiguous from text alone (e.g., \textit{Emphasize, Joke, Taunt}) and require nuanced non-verbal cues for correct interpretation.}
    \label{fig:per_class_accuracy}
\end{figure*}

\subsection{Analysis of Modality-Reliant Intents}
To understand the source of our model's robustness, we analyzed its per-class performance on MIntRec2.0, as shown in Figure~\ref{fig:per_class_accuracy}. This analysis reveals a clear and insightful pattern.

While MAG-BERT may perform marginally better on some high-frequency, textually straightforward classes, DyKen-Hyena achieves substantial, and in some cases dramatic, accuracy improvements on intents that are highly reliant on non-verbal context for disambiguation. For example:
\begin{itemize}
    \item \textbf{Joke}: An intent often impossible to detect from literal text, sees a massive accuracy gain. The model learns to associate playful vocal tones and facial expressions with otherwise neutral text.
    \item \textbf{Doubt}, \textbf{Emphasize}, \textbf{Flaunt}, and \textbf{Taunt}: These categories are fundamentally defined by paralinguistic cues such as vocal tone, prosody, and facial expressions. Our model consistently outperforms the baseline by a large margin on all of them.
\end{itemize}
This provides strong qualitative evidence for our central hypothesis. By dynamically generating per-token kernels from audio-visual signals, DyKen-Hyena can modulate the interpretation of text in real-time. It effectively learns to "listen" to the tone while "reading" the words, a capability that late-fusion models inherently lack. This deeper understanding of nuanced, well-formed multimodal intents is the likely reason for its vastly superior ability to detect out-of-scope samples, as it has a much stronger internal model of what constitutes a valid, in-scope intent.

\begin{table*}[!t]
\centering
\small
\begin{adjustbox}{width=\textwidth,center}
\begin{tabular}{l|S[table-format=2.2] S[table-format=2.2] S[table-format=2.2] S[table-format=2.2] S[table-format=2.2] S[table-format=2.2]|S[table-format=2.2] S[table-format=2.2] S[table-format=2.2] S[table-format=2.2]}
\toprule
\textbf{Model Variant} & {\textbf{Acc}} & {\textbf{WF}} & {\textbf{WP}} & {\textbf{F1}} & {\textbf{Prec}} & {\textbf{Rec}} & {\textbf{oid\_acc}} & {\textbf{F1-IS}} & {\textbf{F1-OOS}} & {\textbf{oid\_f1}} \\
\midrule
\textbf{DyKen-Hyena (Full)} & \textbf{60.54} & \textbf{59.79} & \textbf{60.21} & \textbf{55.02} & \textbf{58.35} & \textbf{54.53} & \textbf{45.63} & \textbf{46.21} & \textbf{38.69} & \textbf{45.97} \\

\midrule
w/o-Attention & 60.39 & 59.44 & 59.59 & 54.31 & 56.81 & 53.95 & 44.87 & 45.84 & 34.51 & 45.48 \\
w/o-DynamicShortConv & 60.24 & 58.74 & 59.21 & 52.82 & 57.30 & 52.95 & 42.16 & 44.55 & 22.31 & 43.83 \\
w/o-LongConv & 60.33 & 59.24 & 59.48 & 53.88 & 56.61 & 53.77 & 44.21 & 45.37 & 31.93 & 44.93 \\
\bottomrule
\end{tabular}
\end{adjustbox}
\caption{Component ablation study on the MIntRec2.0 test set (\%). Removing the Dynamic Short Convolution has the most detrimental effect, especially on F1-OOS, highlighting its critical role.}
\label{tab:ablation_study_mintrec2}
\end{table*}

\begin{table*}[!t]
\centering
\small
\begin{tabular}{l|S[table-format=2.2] S[table-format=2.2] S[table-format=2.2] S[table-format=2.2] S[table-format=2.2] S[table-format=2.2]}
\toprule
\textbf{Model Variant} & {\textbf{Acc}} & {\textbf{F1}} & {\textbf{Prec}} & {\textbf{Rec}} & {\textbf{WF}} & {\textbf{WP}} \\
\midrule
\textbf{DyKen-Hyena (Full)} & \textbf{73.66} & \textbf{69.26} & \textbf{70.30} & \textbf{69.54} & \textbf{73.05} & \textbf{73.30} \\

\midrule
w/o-Attention & 72.09 & 67.99 & 68.32 & 68.82 & 71.64 & 72.19 \\
w/o-DynamicShortConv & 72.09 & 67.80 & 67.85 & 68.97 & 72.00 & 72.95 \\
w/o-LongConv & 72.05 & 67.80 & 68.26 & 68.73 & 71.69 & 72.55 \\
\bottomrule
\end{tabular}
\caption{Component ablation study on the MIntRec test set (\%). All components contribute positively to the final performance.}
\label{tab:ablation_study_mintrec1}
\end{table*}

\subsection{Component Ablation Study}
Our ablation study, presented in Tables \ref{tab:ablation_study_mintrec2} and \ref{tab:ablation_study_mintrec1}, systematically validates the contribution of each key component in our architecture. The most critical component is the dynamic short convolution.
\begin{itemize}
    \item \textbf{w/o-DynamicShortConv}: Removing the entire dynamic local processing path by feeding the attention output directly to the long convolution causes the most severe performance degradation. On MIntRec2.0, the crucial F1-OOS metric plummets from 38.69\% to 22.31\%, a drop of 16.38\%. This unequivocally demonstrates that our proposed dynamic modulation mechanism is the central driver of the model's robustness and fusion capabilities.
    \item \textbf{w/o-Attention}: Replacing the cross-modal attention mechanism with simple mean-pooling of the audio-visual features before the kernel generator also hurts performance significantly. This proves that dynamically \textit{attending} to the most relevant non-verbal cues for each token is far superior to simple, uniform aggregation.
    \item \textbf{w/o-LongConv}: Removing the parallel FFT-based long convolution also leads to a noticeable drop in performance. This shows that the dynamic, local, context-aware feature extraction and the static, global, long-range feature extraction are complementary and synergistic.
\end{itemize}

\subsection{Analysis of Dynamic Kernel Size}
We analyzed the impact of the dynamic kernel size ($K_s$) in Tables \ref{tab:kernel_size_mintrec2} and \ref{tab:kernel_size_mintrec1}. On MIntRec2.0, we observe an interesting trade-off: a kernel size of $K_s=1$ achieves the highest overall accuracy, suggesting that direct token-to-frame modulation without local context is effective for in-domain classification. However, $K_s=3$ achieves vastly superior open-set detection results (F1-OOS of 38.69\% vs. 28.76\% for $K_s=1$). This indicates that incorporating a small local textual context (the immediate neighboring tokens) into the dynamic modulation helps the model form a more robust and generalizable understanding of intent, which is crucial for identifying novel or out-of-scope inputs. On the simpler MIntRec dataset, $K_s=3$ was the clear winner across all metrics, making it a robust and well-balanced default choice for our architecture.

We hypothesize this is because $K_s=3$ allows the non-verbal cues to modulate a small local phrase (the token and its immediate neighbors) rather than just an isolated token. This more closely mimics human communication, where a tone of voice or a facial expression often "colors" an entire phrase, not just a single word. This phrasal-level modulation creates a more robust and contextually-aware representation, which is crucial for distinguishing valid intents from novel out-of-scope inputs. The performance decline at $K_s=5$ suggests that a larger kernel may introduce irrelevant context, diluting the precise, local impact of the audio-visual cue and increasing the learning difficulty for the kernel generator. Therefore, $K_s=3$ strikes a balance, capturing essential local context without introducing noise, making it a robust default choice for our architecture.

\begin{table*}[!t]
\centering
\small
\begin{adjustbox}{width=\textwidth,center}
\begin{tabular}{c|S[table-format=2.2] S[table-format=2.2] S[table-format=2.2] S[table-format=2.2] S[table-format=2.2] S[table-format=2.2]|S[table-format=2.2] S[table-format=2.2] S[table-format=2.2] S[table-format=2.2]}
\toprule
\textbf{Kernel Size} & {\textbf{Acc}} & {\textbf{WF}} & {\textbf{WP}} & {\textbf{F1}} & {\textbf{Prec}} & {\textbf{Rec}} & {\textbf{oid\_acc}} & {\textbf{F1-IS}} & {\textbf{F1-OOS}} & {\textbf{oid\_f1}} \\
\midrule
1 & \textbf{61.00} & \textbf{59.99} & \textbf{60.64} & 54.53 & \textbf{58.97} & 54.21 & 43.94 & 46.11 & 28.76 & 45.55 \\
\textbf{3} & 60.54 & 59.79 & 60.21 & \textbf{55.02} & 58.35 & \textbf{54.53} & \textbf{45.63} & \textbf{46.21} & \textbf{38.69} & \textbf{45.97} \\
5 & 60.58 & 59.63 & 60.28 & 54.44 & 57.96 & 54.27 & 44.78 & 45.92 & 33.80 & 45.53 \\

\bottomrule
\end{tabular}
\end{adjustbox}
\caption{Analysis of dynamic kernel size ($K_s$) on MIntRec2.0 (\%). A trade-off is observed between accuracy ($K_s=1$) and OOS robustness ($K_s=3$).}
\label{tab:kernel_size_mintrec2}
\end{table*}

\begin{table*}[!t]
\centering
\small
\begin{tabular}{c|S[table-format=2.2] S[table-format=2.2] S[table-format=2.2] S[table-format=2.2] S[table-format=2.2] S[table-format=2.2]}
\toprule
\textbf{Kernel Size} & {\textbf{Acc}} & {\textbf{F1}} & {\textbf{Prec}} & {\textbf{Rec}} & {\textbf{WF}} & {\textbf{WP}} \\
\midrule
1 & 70.97 & 66.50 & 67.66 & 67.36 & 70.43 & 71.25 \\

\textbf{3} & \textbf{73.66} & \textbf{69.26} & \textbf{70.30} & \textbf{69.54} & \textbf{73.05} & \textbf{73.30} \\

5 & 71.96 & 68.04 & 68.70 & 68.61 & 71.58 & 72.10 \\
\bottomrule
\end{tabular}
\caption{Analysis of dynamic kernel size ($K_s$) on MIntRec (\%). $K_s=3$ provides the best performance across all metrics.}
\label{tab:kernel_size_mintrec1}
\end{table*}

\section{Conclusion}
In this paper, we proposed \textbf{DyKen-Hyena}, a novel architecture for multimodal intent recognition that pioneers a deep, early-stage fusion strategy. By leveraging cross-modal attention to dynamically generate per-token convolutional kernels, our model learns to modulate textual feature extraction based on real-time audio and visual cues. Our extensive experiments on the MIntRec and MIntRec2.0 datasets demonstrate that DyKen-Hyena achieves new state-of-the-art results. Its strength is particularly evident on intents that are highly dependent on non-verbal context and in the challenging task of out-of-scope detection. Ablation studies confirm that the proposed dynamic modulation mechanism is the key driver of this success. This work highlights a promising new direction for multimodal learning, moving beyond high-level feature concatenation towards more deeply integrated and context-aware systems that more closely mimic human multimodal understanding.

\section*{Limitations}
While our model demonstrates strong performance, we identify a few limitations for future work. First, the computational cost is moderately increased due to the attention and dynamic convolution steps, although it remains more efficient than purely attention-based fusion at every layer. Second, our approach currently relies on an explicit feature alignment step; exploring end-to-end models that handle unaligned sequences is a key area for future research. Finally, our Kernel Generator is a simple MLP; investigating more sophisticated architectures for translating multimodal context into processing kernels could yield further improvements.

\section*{Ethics Statement}
This research focuses on the fundamental scientific challenge of improving human-computer interaction and multimodal understanding. We used publicly available and anonymized resources (MIntRec and MIntRec2.0) that were created and shared by the research community for academic purposes. We acknowledge that any technology capable of interpreting nuanced human behavior could potentially be misused. We therefore advocate for the responsible development and deployment of such technologies, with careful consideration for privacy and fairness.

\bibliography{custom}
\bibliographystyle{acl_natbib}

\end{document}